\pdfoutput=1
\documentclass[11pt]{article}

\usepackage{amsmath}
\usepackage{amssymb}
\usepackage{multirow}
\usepackage{tcolorbox}
\usepackage{colortbl}
\usepackage{makecell}
\usepackage{booktabs}
\usepackage[preprint]{acl}

\usepackage{times}
\usepackage{latexsym}

\usepackage[T1]{fontenc}
\usepackage[utf8]{inputenc}

\usepackage{microtype}

\usepackage{inconsolata}

\usepackage{graphicx}

\usepackage{bbm}

\newcommand\matt[1]{{\textcolor{black}{#1}}}
\newcommand\camera[1]{{\textcolor{black}{#1}}}

\title{EmoDynamiX\includegraphics[scale=0.025]{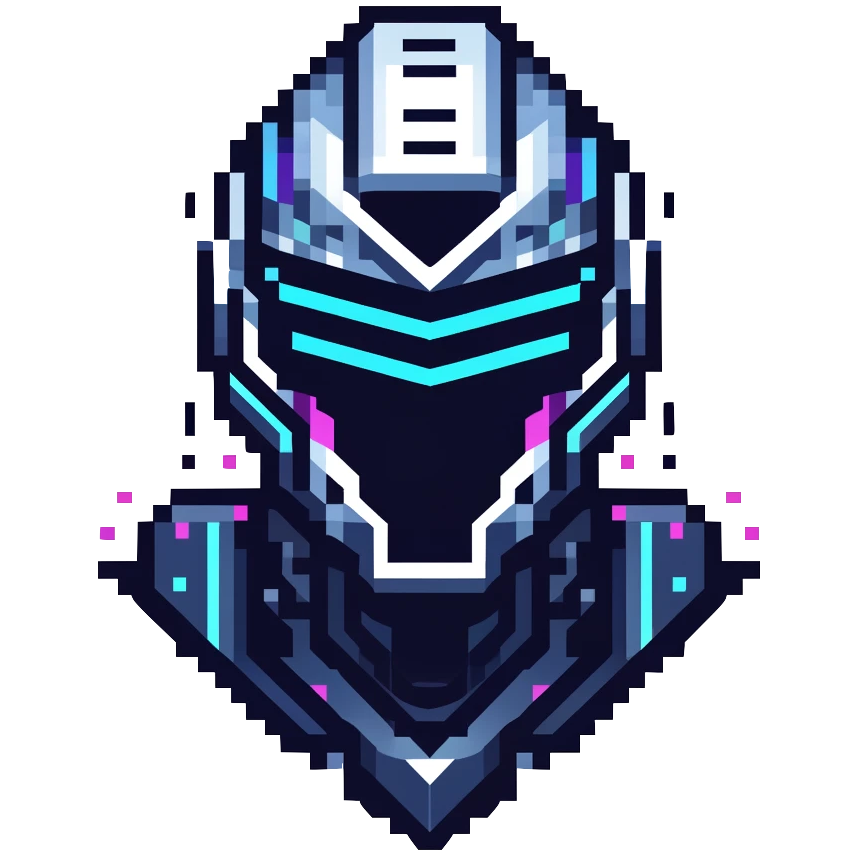}: \underline{Emo}tional Support Dialogue Strategy Prediction \\ by Modelling Mi\underline{X}ed \underline{Emo}tions and Discourse \underline{Dynami}cs}

\author{
  \textbf{Chenwei Wan\textsuperscript{1,2}\thanks{Work done during the first author's internship at Inria.}}\hspace{24pt}%
  \textbf{Matthieu Labeau\textsuperscript{1}}\hspace{24pt}%
  \textbf{Chlo\'e Clavel\textsuperscript{2}}
\\
  \textsuperscript{1}LTCI, T\'el\'ecom Paris, Institut Polytechnique de Paris, France \\
  \textsuperscript{2}Inria Paris, France
\\
  \texttt{\{chenwei.wan, matthieu.labeau\}@telecom-paris.fr} \\
  \texttt{chloe.clavel@inria.fr} \\
}

\begin{document}
\maketitle
\begin{abstract}

Designing emotionally intelligent conversational systems to provide comfort and advice to people experiencing distress is a compelling area of research. Recently, with advancements in large language models (LLMs), end-to-end dialogue agents without explicit strategy prediction steps have become prevalent. However, implicit strategy planning lacks transparency, and recent studies show that LLMs' inherent preference bias towards certain socio-emotional strategies hinders the delivery of high-quality emotional support. To address this challenge, we propose decoupling strategy prediction from language generation, and introduce a novel dialogue strategy prediction framework, \textbf{EmoDynamiX}, which models the discourse dynamics between user fine-grained emotions and system strategies using a heterogeneous graph for better performance and transparency\footnote{Our code is available at \url{https://github.com/cw-wan/EmoDynamiX-v2}.}. Experimental results on two ESC datasets show EmoDynamiX outperforms previous state-of-the-art methods with a significant margin (better proficiency and lower preference bias). Our approach also exhibits better transparency by allowing backtracing of decision making.

% Anonymous repo: https://anonymous.4open.science/r/EmoDynamiX-v2-F6C6
% Github: https://github.com/cw-wan/EmoDynamiX-v2

\end{abstract}

\section{Introduction}

\begin{figure}[t]
\scalebox{0.98}{
  \includegraphics[width=\columnwidth]{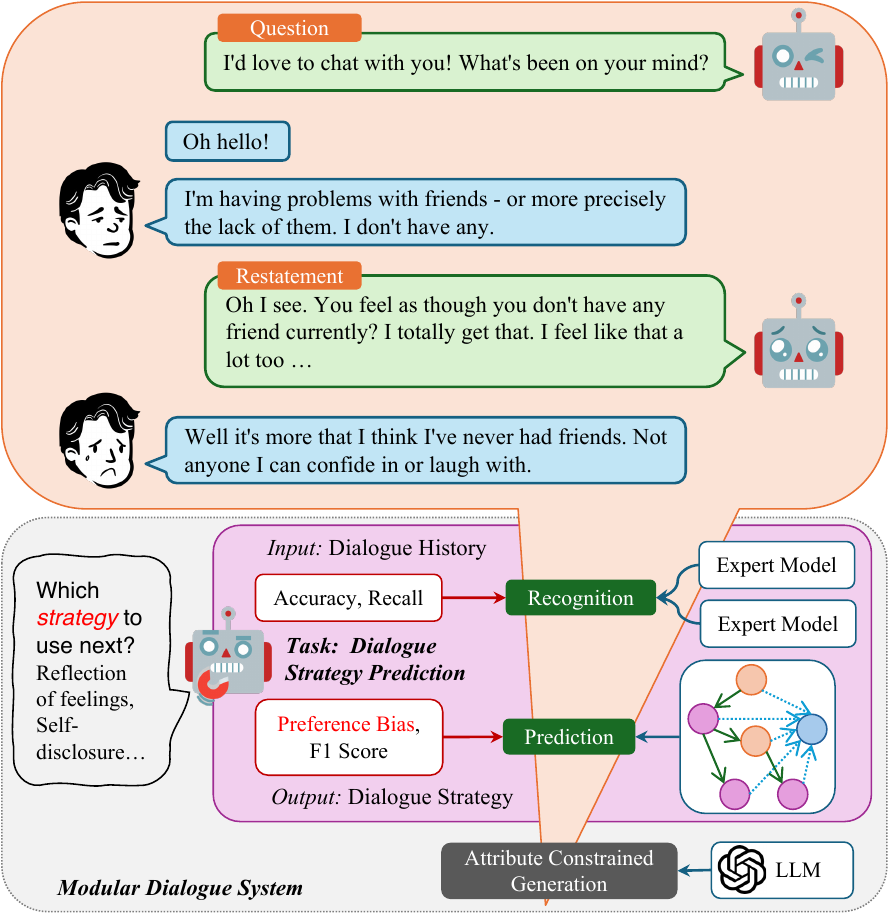}}
  % \caption{This figure shows a possible modular ESC dialogue system, in which \textcolor{Plum}{our dialogue strategy prediction framework} outputs the next dialogue strategy to be used to guide an external generative model.}
  \caption{This figure shows a \textit{possible} modular ESC dialogue system, within which \textcolor{violet}{our dialogue strategy prediction framework} outputs the next dialogue strategy to be used to guide an external generative model.}
  \label{fig:task}
\end{figure}

Providing early intervention for individuals experiencing distress from life challenges is crucial for enabling them to transition toward positive lifestyles and, consequently, fostering a more caring society. This need has inspired the NLP community to develop effective Emotional Support Conversation (ESC) systems \citep{liu-etal-2021-towards}. These systems aim to alleviate the distress of help-seekers and can be seen as a first step in helping them to find healthcare professionals.  Recently, with the release of multi-turn and human-evaluated ESC datasets (\citealp{liu-etal-2021-towards}; \citealp{wu2022anno}), data-driven approaches have begun to surpass rule-based methods (\citealp{van2012bdi}; \citealp{van2012conversation}).% and reinforcement learning-based ones with user simulation \citep{steve2013pomdp}.

Previous work on data-driven ESC has primarily focused on modular dialogue systems as defined by \citet{clavel2022socio}. Modular systems feature a three-fold workflow: recognizing, planning, and generating. Examples include \citet{tu-etal-2022-misc}, \citet{deng-etal-2023-knowledge}, \citet{liu-etal-2021-towards} and \citet{cheng-etal-2022-improving}. In these systems, the socio-emotional strategy is selected based on the recognition of the user's state, and responses are generated using customized language decoders conditioned on the predicted strategy. With the emergence of Large Language Models (LLMs) offering enhanced capabilities, LLMs have increasingly dominated both modular and, particularly, end-to-end ESC systems (\citealp{zheng2023building}), where strategy planning has shifted from an explicit process to a more implicit, hidden mechanism.

However, implicit dialogue strategy planning with LLMs faces two challenges. First, transparency is often lacking in such implicit decision-making processes \camera{due to the well-known "black-box" property of LLMs} (\citealp{ludan-etal-2023-explanation}; \citealp{chhun-etal-2024-language}; \citealp{lu2024struxllmdecisionmakingstructured}). Second, recent studies show that preference biases inherited from pre-training data often cause LLMs to struggle with balancing social-oriented and task-oriented goals. \citet{abulimiti-etal-2023-kind} found that in peer-tutoring dialogues, LLMs like ChatGPT frequently prioritize non-hedging strategies, even in situations where hedging strategies would be more appropriate for repairing low rapport between peers. Similarly, \citet{kang-etal-2024-large} observed that LLMs' strong predisposition towards certain strategies can undermine the outcome of the current stage of ESC. As a result, the overall objectives may be significantly compromised.

To address this limitation, introducing external strategy planners, which offer greater controllability by enabling us to explicitly exclude inappropriate strategies in specific contexts, stands out to be a promising solution. It has been evidenced by both automatic metrics and human evaluations that, \camera{an explicit decision-making module} can more effectively mitigate preference bias, and \camera{improved proficiency in dialogue strategy actually enhances overall generation quality} \citep{kang-etal-2024-large}. \camera{This foundational insight leads us to \textit{isolate and focus specifically on dialogue strategy prediction}—a previously intermediate step—with three primary goals:
(1) better alignment with human expert strategies, (2) reduced preference bias, and (3) improved transparency.}

\camera{Additionally, as more powerful LLMs and improved controlled generation techniques continue to emerge \citep{Dathathri2020Plug}, focusing on dialogue strategy prediction offers a more economical pathway to addressing current dialogue system limitations. Explicit dialogue strategy prediction can serve as a flexible, plug-and-play module compatible with state-of-the-art LLMs or as a foundational component for future RL-based methods, where better alignment with human expert strategies is typically a critical first step \citep{deng2024plugandplay}.}

Therefore, we treat socio-emotional strategy prediction as \textit{an independent task}, as previously explored by \citet{lorraine2023anewtask}. We illustrate the scope of this task in Figure~\ref{fig:task}. From this perspective, we raise the following research questions:

\begin{itemize}
\item \textbf{RQ1}: \camera{Can we %smartly
build a dedicated framework for socio-emotional dialogue strategy prediction that is \textit{more transparent by design}, while outperforming prompting or fine-tuning LLMs in terms of proficiency?}
\item \textbf{RQ2}: Given the importance of emotional intelligence in delivering effective emotional support, can we boost strategy prediction in ESC by by accounting the user's emotion using an ERC (Emotion Recognition in Conversations) module? 
\end{itemize}

In addressing \textbf{RQ1}, we introduce EmoDynamiX, \camera{a decision-making framework} that integrates multiple expert models and incorporates a heterogeneous graph learning module to capture the dynamic interactions between system strategies and user emotions. With graphs, we backtrace the decision-making process, making a step towards greater transparency. We also utilize dummy nodes (\citealp{liu2022boosting}; \citealp{scarselli2008graph}) for role-aware information aggregation, enhancing the overall performance.

For \textbf{RQ2}, we design a mixed-emotion module to effectively integrate ERC into our framework: (1) By using emotion distributions instead of discrete labels, we reduce the risk of error propagation, as there are domain gaps between ERC and ESC datasets. (2) By tuning emotion distributions, we can effectively model nuanced emotion categories by fusing primary emotions.

We validate the effectiveness of our proposed framework through comparative experiments on two public ESC datasets. The results demonstrate that EmoDynamiX significantly outperforms all previous baselines, achieving superior F1 scores and a notable reduction in preference bias. 

\section{Related Work}

\subsection{Emotional Support Conversation}

 The goal of ESC is both social-oriented and task-oriented. It aims to alleviate distress by expressing empathy and providing suggestions \cite{liu-etal-2021-towards,cheng-etal-2022-improving}. %\citet{liu-etal-2021-towards} introduced a joint-learning baseline to condition response generation with predicted strategies. \citet{cheng-etal-2023-pal} infer the user's persona to create more personalized ESC systems. \citet{cheng-etal-2022-improving}  and \citet{LI2024111201} focus on identifying strategies that optimize future user feedback.
%, as demonstrated by \citet{cheng-etal-2022-improving} and \citet{LI2024111201}.
Modelling user states is thus a critical topic in ESC. Previous work commonly approaches this by querying commonsense knowledge graphs \citep{tu-etal-2022-misc, deng-etal-2023-knowledge, ijcai2022p0600, zhao-etal-2023-transesc, LI2024111201}. These queries are constructed by concatenating the current utterance with specific knowledge relations, such as \texttt{xReact}. The queries are then fed into COMET \citep{hwang2021comet}, a generative model pre-trained on commonsense knowledge graphs, which returns the user's emotional reaction (\texttt{xReact}) to the situation. However, commonsense knowledge is too general to capture the fine-grained emotional states. %entailed in specific speaker turns like "Okay". 

In contrast, models specialized in dialogue with context-aware architectures, such as sequential or graph-based models, trained on ERC datasets, can handle these nuances more effectively. Additionally, emotions are frequently mixed in real-life situations, and contradictory emotions (like \textit{Sadness} and \textit{Joy}) could coexist in specific contexts \citep{braniecka2014mixed}. Our mixed-emotion modelling approach handles this complexity better and can model a large set of subtle emotional expressions by combining primary emotions without further human annotations (as demonstrated in Section~\ref{sec:case_study}).

EmoDynamiX features two key distinctions: (1) We provide an alternative to knowledge-based user state modelling: a mixed-emotion module based on label distributions predicted by a pre-trained ERC model (2) While previous works have explored various dialogue graph structures \citep{LI2024111201, ijcai2022p0600, zhao-etal-2023-transesc}, our method incorporates discourse structure, which has been proven effective in various dialogue tasks \citep{chen-yang-2021-structure, Li_Zhu_Mao_Cambria_2023, zhang-etal-2023-dualgats}, but remains underexplored in ESC.

\subsection{Graph Learning in Conversational Tasks}

Graph-based approaches have proven effective in various dialogue-related tasks. In recognition tasks, such as conversational emotion recognition and dialogue act recognition, the target speaker turn aggregates information from its neighbors according to the graph structure. Studies by \citet{ghosal-etal-2019-dialoguegcn}, \citet{ishiwatari-etal-2020-relation}, \citet{wang-etal-2020-integrating}, \citet{10096805}, and \citet{shen-etal-2021-directed} design dialogue graphs based on interactions between speaker roles. \citet{Li_Zhu_Mao_Cambria_2023} and \citet{zhang-etal-2023-dualgats} construct dialogue graphs based on discourse dependencies parsed with a pre-trained expert model, an approach also applied by \citet{chen-yang-2021-structure} and \citet{ijcai2021p0524} in dialogue summarization. \citet{10.1145/3583780.3614758} incorporate commonsense knowledge as heterogeneous nodes. Furthermore, \citet{hu-etal-2021-mmgcn} and \citet{10203083} model multi-modal fusion in dialogue graphs.

In predictive dialogue tasks, such as forcasting the next dialogue act, decisions rely on global information extracted from graphs. Previous works have utilized simple readout functions, such as mean/max pooling \citep{joshi2021dialograph} and linear layers \citep{raut-etal-2023-sentiment}. Our approach introduces dummy nodes as special placeholders for information aggregation. While dummy nodes have been previously used in other graph-learning tasks, such as graph classification and subgraph isomorphism matching~\citep{liu2022boosting}, they have primarily served as alternatives to \textit{readout} functions, \matt{which are used} for obtaining embeddings of graphs or subgraphs. %In our design, however, 
We are the first to employ dummy nodes in a predictive dialogue task, \matt{which is particularly useful,}
%makes sense 
since it allows to clearly model role-aware interactions with previous speaker turns.

\section{Problem Formulation}

The task of predicting the next dialogue strategy can be written as a multi-class classification problem: \matt{assuming a dialogue comprising $T$ speaker turns, we define the dialogue history  as \(H^{T} = \{U^{T}, A^{T}, ST^{T}\}\) where \(U^{T} = \{u_t\}_{t=1}^{T}\) is the sequence of utterances, and each \(u_t = \{w_n\}_{n=1}^{N^t}\) is a sequence of \(N^t\) words. \(A^{T} = \{a_t\}_{t=1}^{T}\)is the sequence of speaker roles, with \(a_t \in \{\texttt{user}, \texttt{system}\}\).
$ST^{T} = \{{st}_t\}_{t=1}^{T}$ is the sequence of possible strategies, but they only exist for the agent. Noting $\mathcal{S}$ the set of strategies, $I_{\texttt{user}} = \left\{t,  a_t = \texttt{user}\right\}$ the indexes of turns where the user is speaking, and denoting similarly $I_{\texttt{agent}}$, we have that $\forall t \in I_{\texttt{agent}}, {st}_t \in \mathcal{S}$ and $\forall t \in I_{\texttt{user}}, {st}_t = \varnothing$.}
Our task is, given a fixed window size of $N-1$, to predict the strategy for agent speaker turns, which we formulate as estimating the probability distribution $\mathbb{P}(st_{N} \mid H_{1}^{N-1})$ upon $\mathcal{S}$ when $t \in I_{\texttt{agent}}$. For the sake of simplicity, from now on we will use indexing from the beginning of the context window and no longer from the entire conversation.

%In a dialogue comprising \matt{$T$} \(t-1\) speaker turns, the dialogue history can be represented as \(H = \{U, A, M, ST\}\). Here, \(U = \{u_i\}_{i=1}^{t-1}\) denotes the sequence of historical utterances, where each utterance \(u_i = \{w_n\}_{n=1}^{N^i}\) is a sequence of \(N^i\) words. \(A = \{a_i\}_{i=1}^{t-1}\), where \(a_i \in \{\texttt{user}, \texttt{system}\}\), indicates the sequence of speaker roles. \( M = \{ m_i \mid m_i = 1 \text{ if } a_i = \texttt{system} \text{ else } 0 \}_{i=1}^{t-1} \) is the mask sequence. Since in the ERC task, the conversational agent is the one that applies socio-emotional strategies with intention, the historical strategy sequence contains only those applied by the agent: \(ST = \{st_j\}_{j=1}^{\sum M}\). Our goal is to predict the next strategy \(st_{\sum M + 1}\) for the conversational agent. This can be formulated as estimating the probability distribution \(P(st_{\sum M + 1} \mid H)\).

\section{Methodology}

\begin{figure*}[t]
  \includegraphics[width=\linewidth]{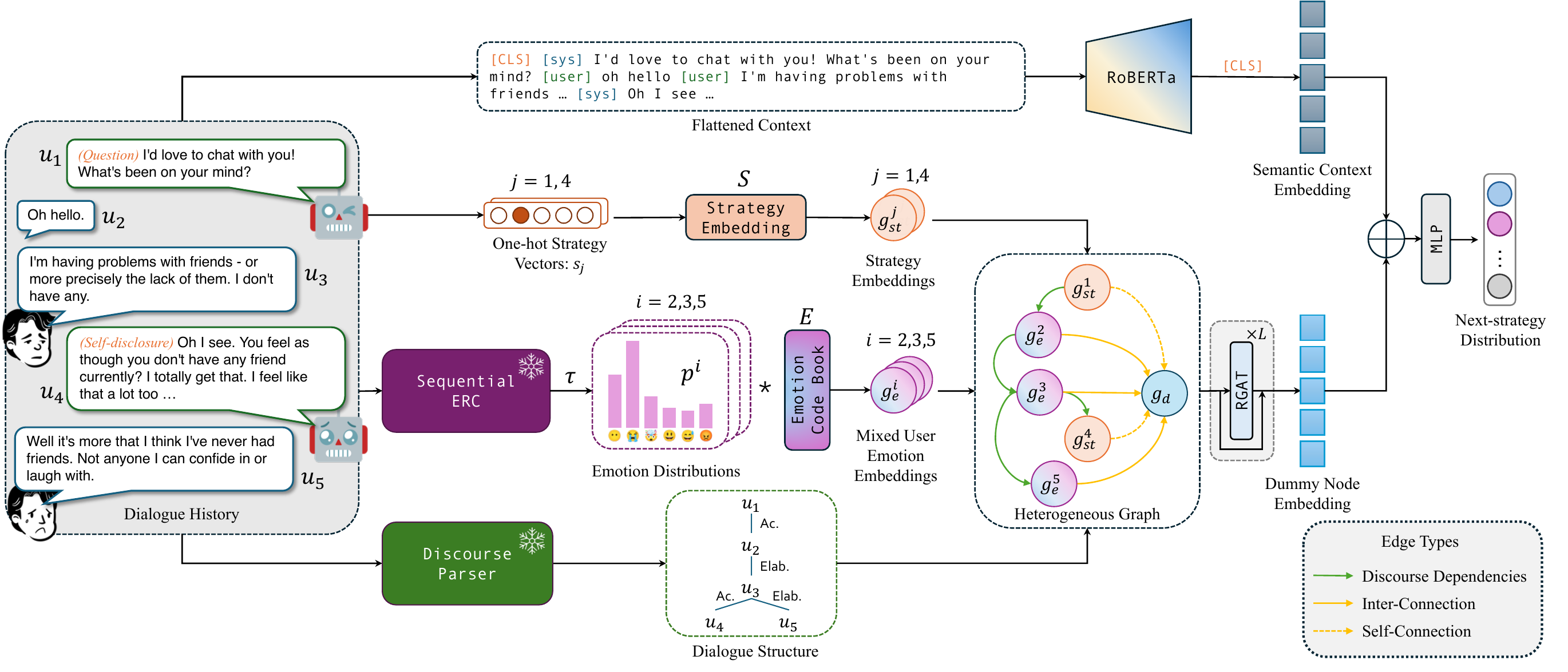}
  \caption{The overview of our proposed model that consists of a semantic modelling module, a heterogeneous graph learning module, and an MLP classification head.}
  \label{fig:overview}
\end{figure*}

Our framework (see Figure~\ref{fig:overview}) comprises three main components: (1) a semantic modelling module for capturing the semantics of the dialogue context; (2) a heterogeneous graph learning module, designed to capture the complex interplay between the user's emotions and system strategies within the dialogue history; and (3) an MLP classification head which integrates the features obtained from the previous modules to produce the prediction result.

\subsection{Semantic Modelling}

To effectively capture the global semantic information in the dialogue history, we adopt a common method of representing the context in a flattened sequence format:
%\begin{equation}
%  \label{eq:context}
%  \texttt{<context>} = \texttt{[}a_1\texttt{]}, u_1, \texttt{[}a_2\texttt{]}, u_2, ...
%\end{equation}
\begin{equation}
  \label{eq:context}
  %\resizebox{!}{0.72em}{
  \texttt{<context>} = \texttt{[}a_{1}\texttt{]}, u_{1}, \texttt{[}a_{2}\texttt{]}, u_{2}, ...
  %}
\end{equation}
For each speaker turn, we use its speaker role as the separating token indicating the start of the corresponding utterance. This sequence is encoded using RoBERTa \citep{liu2019roberta}:
\matt{\begin{equation}
\label{eq:semantic_encode}
  \boldsymbol{C} = \texttt{Roberta}(\texttt{[CLS]}, \texttt{<context>})
\end{equation}}
We use the embedding of the \texttt{[CLS]} token \matt{$\boldsymbol{C}_{\texttt{[CLS]}}$} from the last hidden layer output as the global semantic representation of the \matt{dialogue history $H_{1}^{N-1}$}. %entire dialogue history.

\subsection{Heterogeneous Graph Learning}

We propose to use a heterogeneous graph (HG) to model the interaction dynamics \matt{in the history $H_{1}^{N-1}$}.
%between the user's history emotional states and history strategies adopted by the conversational agent. 
\matt{This graph is formally defined as $\mathcal{G} = \{\mathcal{V}, \mathcal{B}\}$, which are respectively the set of nodes and edges. Both can be of several \textit{types}: broadly, \textit{node types} correspond to past strategies adopted by the conversational agent, the user's previous emotional states, and the strategy to be predicted; while \textit{edge types} model the discourse dependencies between dialogue turns and facilitate information aggregation.}

%This graph is formally defined as $\mathcal{G} = \{\mathcal{V}, \mathcal{B}, \mathcal{N}, \mathcal{R}\}$, where $\mathcal{V}$ represents nodes, $\mathcal{B}$ denotes edges, $\mathcal{N}$ specifies node types, and $\mathcal{R}$ indicates edge types. Each node $v \in \mathcal{V}$ or edge $b \in \mathcal{B}$ is projected to its type via a mapping function: $\phi(v): \mathcal{V} \rightarrow \mathcal{N}$, $\psi(b): \mathcal{B} \rightarrow \mathcal{R}$, respectively.

\noindent \matt{\textbf{Node Types:} Our heterogeneous graph contains $N$ nodes $\mathcal{V} = \{v_i\}_{i=1}^N$, each corresponding to a speaker turn, including the target turn $N$. For user turns $i \in I_{\texttt{user}}$, $v_i$ is an \textit{emotion node}, which encapsulates the fine-grained emotion state of the user. %, the construction of which is detailed in Section \ref{sec:mixederc}. 
For agent turns $i \in I_{\texttt{agent}}$, $v_i$ is a \textit{system strategy node}, %taking the form of a one-hot vector 
that represents the specific conversational strategy implemented by the agent. Lastly, we introduce the \textit{dummy node} $v_N$ as a placeholder for the target utterance, aggregating information from the two other node types and their interactions.}

%\noindent\textbf{Node Types:} In our heterogeneous graph, each node corresponds to a dialogue turn and is classified into one of three types: user emotion node, system strategy node, and dummy node. The user emotion nodes, symbolized by $e$, encapsulate the fine-grained emotion states of the user from turn to turn. Our method used to obtain such information is described in detail at section \ref{sec:mixederc}. System strategy nodes, indicated by $s$, are one-hot vectors that represent the specific conversational strategies implemented by the agent as responses to the user. Due to the predictive nature of our task, where the content of the target utterance is unknown, we introduce the dummy node, denoted by $d$. This node acts as a placeholder for the target utterance, aggregating information from the previous two node types and their interactions. Each dialogue graph contains one single dummy node. Collectively, the set of node types, $\mathcal{N}$, is defined as $\{e, s, d\}$.

\noindent\textbf{Edge Types:} The edges in our heterogeneous graph fulfill dual roles: they model the discourse dependencies between dialogue turns and facilitate information aggregation towards the dummy node. 
\matt{Discourse dependencies correspond to edges between and within emotion and system strategy nodes: we follow~\citet{asher-etal-2016-discourse}'s definition (which include categories such as \textit{Comment} and \textit{Elaboration}) and pre-train a discourse parser as proposed by \citet{chi-rudnicky-2022-structured} on the multi-party discourse dataset STAC \citep{asher-etal-2016-discourse}. Details regarding the training procedure can be found in Appendix~\ref{sec:append_impl_sub_modules}. We note $\mathcal{R}_{\text{Discourse}}$ the set of possible dependencies; then, $\forall (i,j)$ such that $1 \leq i, j \leq N-1, \langle v_i, v_j\rangle \in \mathcal{R}_{\text{Discourse}}$.
We give more details on these dependencies in Appendix~\ref{sec:appendix_datasets_expert}.
The remaining edges are aggregating information from system strategy nodes, which we call \textit{self-reference}: $\forall i \in I_{\texttt{agent}}, \langle v_i, v_N\rangle = r_{\text{self}}$; and from user emotion nodes, which we call \textit{inter-reference}: $\forall i \in I_{\texttt{user}}, \langle v_i, v_N\rangle = r_{\text{inter}}$. We note $\mathcal{R} = \mathcal{R}_{\text{Discourse}} \cup \{r_{\text{self}}, r_{\text{inter}}\} $ the set of edge types.}

%Firstly, for discourse dependencies, we follow \citet{asher-etal-2016-discourse}'s definition of discourse dependencies, and pre-train a discourse parser as proposed by \citet{chi-rudnicky-2022-structured} on the multi-party discourse dataset STAC \citep{asher-etal-2016-discourse}. Our discourse parser achieves a 59.0 F1 score on link and relation predictions, comparable to the state-of-the-art results. This parser is used to identify edges and their types between user emotion nodes and system strategy nodes within the ESConv dialogues. We denote all 16 discourse dependency categories ( such as \textit{Comment} and \textit{Elaboration}, more to be found in Appendix~\ref{sec:appendix_datasets}) as the set $\mathcal{R}^{Dis}$. For each edge $b=\langle v_1, v_2\rangle$ that $\psi(b) \in \mathcal{R}^{Dis}$, we have $\phi(v_1), \phi(v_2) \in \{e, s\}$. Regarding the information aggregation for the dummy node, we introduce two relation types: $r^{Self}$ for aggregating information from system strategy nodes (self-reference) and $r^{Inter}$ for aggregating from user emotion nodes (inter-reference). Formally, for each edge $b=\langle v_1, v_2\rangle$ where $\psi(b) = r^{Self}$, we have $\phi(v_1) = s$ and $\phi(v_2) = d$. Similarly, for edges where $\psi(b) = r^{Inter}$, we have $\phi(v_1) = e$ and $\phi(v_2) = d$. Consequently, the set of edge relations can be expressed as $\mathcal{R} = \mathcal{R}^{Dis} \cup \{r^{Self}, r^{Inter}\}$.

\matt{We will describe next how we obtain node embeddings for these three node types, and how the different edge types affect their aggregation.}

\subsubsection{User Emotion Node Embedding: Mixed Emotion Method}
\label{sec:mixederc}

Unlike previous work that relies on commonsense knowledge, we propose leveraging a pre-trained ERC model to predict emotion distributions from user utterances. We then utilize the knowledge contained in these distributions to create embeddings for fine-grained user states using a mixed-prototype approach. 

\paragraph{Training an ERC model} Our emotion recognition model consists of a RoBERTa encoder with an MLP classifier. To incorporate the global dialogue context while classifying individual utterances, we concatenate all utterances into a single sequence. \matt{We train our model with} the DailyDialog dataset~\citep{li-etal-2017-dailydialog}; since it does not provide annotations for speaker roles, we use the special token \texttt{[SEP]} as the common delimiter between all utterances:
\matt{\begin{equation}
  \label{eq:ercontext}
  \texttt{<ucontext>} = \texttt{[SEP]}_{1}, u_{1}, \texttt{[SEP]}_{2}, u_{2}, ...
\end{equation}
}
This concatenated sequence is then encoded by RoBERTa, from which we extract the embeddings of the $\texttt{[SEP]}_i$ tokens (preceding each utterance) from the last hidden layer. These embeddings serve as representations for the corresponding utterances. We note $\mathcal{E}$ the set of emotions; then, the embeddings are fed into an MLP to derive a vector $\boldsymbol{z}^{i} \in \mathbb{R}^{|\mathcal{E}|}$ of scores for each utterance $u_i$:
\matt{\begin{equation}
  \label{eq:erccls}
  \begin{gathered}
  \boldsymbol{C}^{ERC} = \texttt{Roberta}(\texttt{[CLS]}, \texttt{<ucontext>}) \\
  \boldsymbol{z}^{i} = \texttt{MLP}(\boldsymbol{C}^{ERC}_{\texttt{[SEP]}_i})
  \end{gathered}
\end{equation}}
For DailyDialog, we categorize emotions into seven groups: Ekman’s six basic emotions plus \textit{Neutral}, collectively referred to as $\mathcal{E}$. Additional statistical details about DailyDialog, along with our motivation for selecting it, are available in Appendix~\ref{sec:appendix_datasets_expert}. Detailed information regarding the implementation and training hyperparameters can be found in Appendix~\ref{sec:append_impl_sub_modules}.

\paragraph{Mixed-emotion module:} %We now introduce the mixed-emotion module designed for our strategy prediction framework.
\matt{To model the user's emotional states, we employ a trainable \textit{emotion codebook}. It takes the form of a parameter matrix $\boldsymbol{E} \in \mathbb{R}^{|\mathcal{E}| \times h}$ with $h$ the embedding size for our heterogeneous graph; each of the $|\mathcal{E}|$ vector $\{ \boldsymbol{E}_k \}_{k=1}^{|\mathcal{E}|}$ encodes a distinct emotion. For an emotion node $v_i$, they are combined into a node embedding $\boldsymbol{g}_e^i$ using the adjusted emotion distribution:
\begin{equation}
  \label{eq:emoembed}
  \boldsymbol{g}_e^{i} = \boldsymbol{p}^{i} \cdot \boldsymbol{E}
\end{equation} 
This distribution is directly obtained through the output scores of our ERC model:
\begin{equation}
  \label{eq:softerc}
  \boldsymbol{p}^{i} = \left[ \frac{\exp(z^{i}_j /\tau)}{\sum_{k}\exp(z^{i}_k/\tau)}
  \right]_{j=1}^{|\mathcal{E}|}
\end{equation}
To utilize the information in the emotion label distribution more effectively, we employ a learnable temperature parameter $\tau$; details on the impact of the initialization of \(\tau\) can be found in Appendix~\ref{sec:appendix_temperature}.}

%To utilize the information in the emotion label distribution effectively, we apply a learnable parameter $\tau$ which will help modulate the distribution:
%\begin{equation}
%  \label{eq:softerc}
%  p_i = \frac{\exp(z_i /\tau)}{\sum_{j}\exp(z_j /\tau)}
%\end{equation}
%Details on the impact of the initialization of \(\tau\) can be found in Appendix~\ref{sec:appendix_temperature}. To model the user's emotional states, we employ a trainable emotion codebook $E \in \mathbb{R}^{m \times h}$, where $m$ represents the number of emotion labels and $h$ the embedding size for our heterogeneous graph. The graph representation of the user’s emotional state $g_e$ is computed by weighting the codebook $E$ with the adjusted label distribution $p$:

%\begin{equation}
%  \label{eq:emoembed}
%  g_v^{(i)} = p^{(i)} \cdot E
%\end{equation}

Our mixed emotion approach draws inspiration from MISC, where a mixed-strategy module is proposed to condition the response generation \citep{tu-etal-2022-misc}, yet %it features two key distinctions: (1) while MISC employs a mixed-strategy method to support its customized decoder, our mixed-emotion technique aims at enhancing strategy planning and is fully compatible with contemporary large language model (LLM) decoders; (2) 
MISC does not study the tuning of the label distribution $\boldsymbol{p}$:
%, which can impact the overall model effectiveness. 
for example, in an ERC dataset where the label "Neutral" is prevalent, refining the distribution to become a little "sharper" could \matt{greatly} mitigate the ambiguity in the model's predictions, especially for underrepresented classes.

\subsubsection{System Strategy Node Embedding}
\label{sec:stembedding}

\matt{For a speaker turn $i \in I_{\texttt{agent}}$, the dialogue strategy information is encoded as a one hot vector $\boldsymbol{s}^{i} \in \{0, 1\}^{|\mathcal{S}|}$. Strategies themselves, as emotions, are represented through a parameter matrix $\boldsymbol{S} \in \mathbb{R}^{|\mathcal{S}| \times h}$. Then, we simply obtain:
\begin{equation}
  \label{eq:stembed}
\boldsymbol{g}^i_{st} = \boldsymbol{s}^{i} \cdot \boldsymbol{S} 
\end{equation}
as the embedding of the system strategy node $v_i$.}
%Dialogue strategies are encoded as one-hot vectors: To obtain graph embeddings for the strategy nodes, we employ an embedding layer as described by the following equation:
%\begin{equation}
%  \label{eq:stembed}
%  g_s^{(i)} = W_{st}s^{(i)} + b_{st}^{(i)}
%\end{equation}
\subsubsection{Dummy Node Embedding}
\label{sec:dummyembedding}

Previous work relies on aggregating heterogeneous graph information using simple readout functions and linear layers, which do not consider speaker roles and lack transparency regarding the contribution of each node to the final decision. To address this, we propose using the dummy node $v_t$ as a placeholder for the target of prediction, which interacts with previous speaker turns in a role-aware manner. \matt{We set the embedding of the dummy node as a parameter vector $\boldsymbol{g}_d \in \mathbb{R}^h$, hence being trainable and shared among all dialog graphs.}

%The embedding of the dummy node, represented as \( g_d \), is initialized with a random vector of dimension \( h \) and shared across all dialogue graphs. We designate this vector as trainable to enhance the graph learning process.

\subsubsection{Relational Graph Attention Layers}

%Given the heterogeneous dialogue graph \(\mathcal{G}\) and the node embeddings \(G = \{g_e^{(i)}\}_{i=1}^{\left \vert \mathcal{V}_e \right \vert} \cup \{g_s^{(j)}\}_{j=1}^{\left \vert \mathcal{V}_s\right \vert} \cup \{g_d\}\), we employ relational graph attention networks (\citealp{busbridge2019relational}) to update the node representations. The relational graph attention (RGAT) layer operates as follows:
\matt{We previously defined the initial node representations for our three node types. Then,} we can employ relational graph attention networks (\citealp{busbridge2019relational}) to update these node representations. \matt{Choosing a number $K$ of attention heads, we define a relation graph attention (RGAT) layer by defining \textit{keys}, \textit{queries} and \textit{values} parameter matrices $\boldsymbol{W}_K^{(r,k)}, \boldsymbol{W}_Q^{(r,k)}, \boldsymbol{W}_V^{(r,k)}$ for each attention head $k$ and possible type of edge $r \in \mathcal{R}$. We begin by computing the attention weights $\alpha_{i,j}^{(r_{ij}, k)}$ between $v_i$ and $v_j$ under relation type $r_{ij} = \langle v_i, v_j\rangle $ using:
\begin{equation}
  \label{eq:rgat}
  \begin{gathered}
  a_{i,j}^{(r_{ij}, k)} = \sigma(\boldsymbol{W}_Q^{(r_{ij},k)}\boldsymbol{g}^i + \boldsymbol{W}_K^{(r_{ij},k)}\boldsymbol{g}^j) \\
  \alpha_{i, j}^{(r_{ij}, k)} = \frac{\exp(a_{i,j}^{(r_{ij}, k)})}{\sum_{r\in \mathcal{R}}\sum_{m \in \mathcal{N}_r(i)}\exp(a_{i,m}^{(r, k)})}
  \end{gathered}
\end{equation}} 
where \(\sigma\) denotes the LeakyReLU function and $\mathcal{N}_r(i)$ denotes the set of the indexes of neighbouring nodes of \(v_i\) under the edge type \(r\). %The relational graph attention (RGAT) layer operates as follows:
%\begin{equation}
%  \label{eq:rgat}
%  \begin{gathered}
%  a_{i,j}^{(r, k)} = \sigma(W_Q^{(r,k)}g_i + W_K^{(r,k)}g_j) \\
%  \alpha_{i, j}^{(r, k)} = \frac{\exp(a_{i,j}^{(r, k)})}{\sum_{r'\in \mathcal{R}}\sum_{t\in \mathcal{N}^{(r')}_i}\exp(a_{i,t}^{(r', k)})}
%  \end{gathered}
%\end{equation}
%Here, \(\sigma\) denotes the LeakyReLU function, \(W_Q^{(r,k)}, W_K^{(r,k)}\) are trainable weights. \(\alpha_{i,j}^{(r, k)}\) is the attention weight between \(g_i\) and \(g_j\) under relation type \(r\). \(\mathcal{N}_i^{(r)}\) denotes the set of neighbouring nodes of \(g_i\) under the edge type \(r\). The multi-head attention can be denoted as:
%\begin{equation}
%  \label{eq:rgatatention}
%  u_i = \Vert_{k=1}^K \sigma(\sum_{r\in\mathcal{R}}\sum_{j\in \mathcal{N}_i^{(r)}}\alpha_{i,j}^{(r,k)}W_V^{(r,k)}g_j)
%\end{equation}
\matt{The result of the multi-head attention for node $v_i$ is then:
\begin{equation}
  \label{eq:rgatatention}
  \boldsymbol{h}^i = \Vert_{k=1}^K \sigma(\sum_{r\in\mathcal{R}}\sum_{m \in \mathcal{N}_r(i)}\alpha_{i,m}^{(r,k)}\boldsymbol{W}_V^{(r,k)}\boldsymbol{g}^m)
\end{equation}}
where \(\Vert\) denotes concatenation. To avoid gradient vanishing, we also add residual connections between RGAT layers and obtain the new representation for node $v_i$:
\begin{equation}
  \label{eq:res}
  {\boldsymbol{g}}^{(1), i} = \boldsymbol{h}^i + \boldsymbol{g}^i
\end{equation}
\matt{In our model,} %Finally, 
we use the embedding \matt{$\boldsymbol{g}^{(L),N}$} of the dummy node after \matt{applying }\(L\) RGAT layers as representation for \textit{the entire heterogeneous dialogue graph}. %This embedding is denoted as \(h_{dummy}^{(L)}\).

\subsection{Next Dialogue Strategy Prediction}
\label{sec:training}

\matt{We concatenate our global semantic embedding $\boldsymbol{C}_{\texttt{[CLS]}}$ with the heterogeneous graph embedding $\boldsymbol{g}^{(L)}_N$; this combined representation is fed into a simple MLP classification layer to compute a probability distribution upon $\mathcal{S}$: 
\begin{equation}
  \label{eq:mlp}
  \boldsymbol{o} = {softmax}(\texttt{MLP}( \boldsymbol{C}_{\texttt{[CLS]}} \Vert \boldsymbol{g}^{(L),N}))
\end{equation}}
%We derive the final representation for predicting the next dialogue strategy by concatenating the global semantic embedding \(C_0\) with the heterogeneous graph embedding \(H_{dummy}^L\). This combined representation is fed into a simple MLP classification head to compute the probability distribution for the next dialogue strategy \(p_{st}\):
%\begin{equation}
%  \label{eq:mlp}
%  p_{st} = \texttt{MLP}(C_0\Vert H_{dummy}^{(L)})
%\end{equation}
which finally gives $\mathbb{P}(st_{N} \mid H_{1}^{N-1}) = {o}_{st_{N}}$. We adopt the \textit{weighted} cross-entropy loss as our training objective: \matt{to address class imbalance, we adjust the loss contribution from each class based on its prevalence, weighting it in proportion inverse to its frequency in the training dataset.}
%The weights for the classes, denoted as \(\{w_{st}^{(i)}\}_{i=1}^{\left\vert\mathcal{P}\right\vert}\), are set in inverse proportion to their frequency in the training dataset. This approach helps address class imbalance by adjusting the loss contribution from each class based on its prevalence:
%\begin{equation}
%  \label{eq:loss}
%  L = -\sum_{i=1}^{\left\vert \mathcal{P} \right\vert} w_{st}^{(i)}\log\frac{\exp(p_{st}^{(i)})}{\sum_{j=1}^{\left\vert \mathcal{P}\right\vert}\exp(p_{st}^{(j)})}y^{(i)}
%\end{equation}

\begin{table*}
  \centering
  \scalebox{0.78}{
\begin{tabular}{clcccccc}
\Xhline{1.25pt}
 & \multirow{2}{*}{\textbf{Model}} & \multicolumn{3}{c}{ESConv} & \multicolumn{3}{c}{AnnoMI} \\
\cline{3-8}
 & & \(\mathcal{M}\)-F1$\uparrow$ & \(\mathcal{W}\)-F1$\uparrow$ & \(\mathcal{B}\)$\downarrow$ & \(\mathcal{M}\)-F1$\uparrow$ & \(\mathcal{W}\)-F1$\uparrow$ & \(\mathcal{B}\)$\downarrow$ \\
\Xhline{1.25pt}
    \multirow{6}{*}{\makecell{Prompting\\ LLMs}} & LLaMA3-70B \citep{llama3} & 15.36 & 18.45 & 1.03 & 8.38 & 9.01 & 1.24 \\
    & \textit{+ 2 shot} & 17.70 & 21.47 & 1.29 & 9.52 & 9.13 & 1.11 \\
    & \textit{+ ERC} & 15.70 & 19.32 & 1.12 & 8.28 & 9.02 & 1.30 \\
    & ChatGPT \citep{chatgpt} & 18.14  &  20.27 & 0.88 & 20.31 & 18.12 & 1.21 \\
    & \textit{+ 2 shot} & 16.55 & 20.01 & 0.73 & 15.29 & 14.22 & 1.39 \\
    & \textit{+ ERC} & 16.50 & 18.79 & 0.77 & 16.17 & 15.60 & 0.89 \\
    \hline
    \multirow{3}{*}{\makecell{Finetuning\\ LLMs}} & RoBERTa \cite{liu2019roberta} & 25.04 & 27.94 & 0.68 & 22.26 & 27.25 & \underline{0.64} \\
    & BART \citep{lewis-etal-2020-bart} & 25.66 & 29.08 & 0.64 & 22.94 & 29.68 & 1.07 \\
    & LLaMA3-8B \citep{llama3} & 25.91 & 29.82 & 0.83 & \underline{23.77} & \underline{29.98} & 0.81 \\
    \hline
    \multirow{5}{*}{\makecell{Specialized\\ Models}} & MISC \citep{tu-etal-2022-misc} & 20.91 &  24.93 & 0.89 & - & - & - \\
    & MultiESC \citep{cheng-etal-2022-improving} & 25.73 & 29.31 & \underline{0.61} & - & - & - \\
    & KEMI \citep{deng-etal-2023-knowledge} & 24.69 &  26.80  & 0.86 & - & - & - \\
    & TransESC \citep{zhao-etal-2023-transesc} & \underline{26.28} &  \underline{31.33}  & 0.73 & - & - & - \\
    \hline
    \rowcolor{gray!20}
    Ours & \textit{\textbf{EmoDynamiX}} & \textbf{27.70}$^\text{†}$ & \textbf{32.71}$^\text{†}$ & \textbf{0.45}$^\text{†}$ & \textbf{27.92}$^\text{†}$ & \textbf{35.33}$^\text{†}$ & \textbf{0.50}$^\text{†}$\\
\Xhline{1.25pt}
  \end{tabular}}
  \caption{\label{result-main}
    Experimental results on two ESC datasets. The best results are \textbf{bolded} and the second best are \underline{underlined}. † indicates statistically significant improvement ($p<0.05$). Since MultiESC has adopted a new set of labels, we merge the updated ones with the original annotations to ensure a fair comparison. Due to the unavailability of TransESC's data preprocessing pipeline, we report the reproduced results based on their released train/dev/test split.
  }
\end{table*}

\section{Experiments}

\subsection{Experimental Setups}
\label{sec:implementation}

\noindent\textbf{Datasets  } To make our model learn strategies beneficial for both social and task-oriented goals, we select two ESC datasets (in English) where dialogues have been human-evaluated and filtered to ensure that the conversational outcomes are positive and the strategies applied are socially appropriate: (i) \textbf{ESConv} \citep{liu-etal-2021-towards}, an ESC dataset annotated by trained crowd-workers. It comprises 1,300 dialogues and features 8 dialogue strategies. We adhere to the official train/dev/test split\footnote{https://huggingface.co/datasets/thu-coai/esconv}.; (ii) \textbf{AnnoMI} \citep{wu2022anno}, an expert-annotated counselling dataset. It includes 133 dialogues and features 9 therapist strategies. We make the train/dev/test split with an 8:1:1 ratio. Detailed dataset statistics are provided in Appendix~\ref{sec:appendix_datasets_esc}. For both datasets, we set the context window size to 5 utterances\footnote{\camera{We employ a \textit{sliding window} method to generate training samples: for each supporter utterance in a dialogue, we take the five preceding utterances as the dialogue history and use the current dialogue strategy as the target for prediction.}}, resulting in 18,376 samples for ESConv and 4,442 samples for AnnoMI.

\noindent\textbf{Baselines  } To provide extensive comparisons, we choose baselines from three criteria: (i) prompting LLMs SOTA in dialogue tasks, using task description and supplementary information: \textbf{ChatGPT}\footnote{\camera{The version of ChatGPT is \texttt{gpt-3.5-turbo-0301}}} \citep{chatgpt} and \textbf{LLaMA3-70B} \citep{llama3} with 2-shot learning (\textit{+2 shot}) or emotion labels (\textit{+ ERC}). We excluded Chain-of-Thought \citep{wei2022chain} prompting because it has already been shown to be ineffective for our task \citep{kang-etal-2024-large}; (ii) fine-tuning LLMs as general-purpose dialogue strategy predictors: \textbf{RoBERTa} \citep{liu2019roberta}, \textbf{BART} \citep{lewis-etal-2020-bart} and \textbf{LLaMA3-8B} \citep{llama3}; (iii) specialized models for emotional support dialogue strategy prediction: \textbf{MISC} \citep{tu-etal-2022-misc}, \textbf{MultiESC} \citep{cheng-etal-2022-improving}, \textbf{KEMI} \citep{deng-etal-2023-knowledge} and \textbf{TransESC} \citep{zhao-etal-2023-transesc}. For more details see Appendix~\ref{sec:baselines}.

\noindent\textbf{Evaluation Metrics} We use macro F1 score (\textbf{\(\mathcal{M}\)-F1}) and weighted F1 score (\textbf{\(\mathcal{W}\)-F1}) as metrics for evaluating the proficiency of strategy prediction models, since the ground truth strategies in the two datasets have been validated by human evaluators. Given the unbalanced nature of ESC datasets, the accuracy score is not an ideal choice as it can unfairly favor models that predominantly predict the majority classes. We also incorporate the preference bias score (\textbf{\(\mathcal{B}\)}) as defined by \citet{kang-etal-2024-large} to quantify the extent to which the model favors its preferred strategies over non-preferred ones (implementation details in Appendix~\ref{sec:preference_bias}). An ideal dialogue strategy predictor should achieve strong F1 scores while minimizing preference bias.

\noindent\textbf{Implementation  Details  } We implemented our proposed method using PyTorch \citep{paszke2019pytorch}, initializing with the pre-trained weights from RoBERTa and employing the tokenization tools from Huggingface Transformers \citep{wolf-etal-2020-transformers}. For optimization, we used the AdamW optimizer \citep{loshchilov2018decoupled}. Detailed hyperparameter settings can be found in Appendix~\ref{sec:hyps}.

\subsection{Overall Performance}
\label{sec:performance}

As shown in Table~\ref{result-main}, EmoDynamiX outperforms previous SOTA methods across all evaluation metrics. Compared to TransESC, EmoDynamiX significantly reduces the preference bias score by 38\% and achieves higher F1 scores. This suggests that while TransESC also models dialogue state transitions, our ERC-based mixed-emotion approach captures the nuances of user emotion states more effectively, leading to better predictions, as further validated by our ablation study. Furthermore, compared to MultiESC, the previous SOTA model with a low bias score, EmoDynamiX excels across all metrics by a substantial margin. Another interesting result is that models based on specific case knowledge (KEMI) and general commonsense knowledge (MISC) are less effective as dialogue strategy predictors.

Comparisons with two LLM prompting baselines indicate that using LLMs alone for emotional support dialogue prediction is significantly constrained by their inherent biases, even when examples or emotion recognition is provided through prompts. The bias scores for LLM-prompting baselines are significantly higher, ranging from 0.77 to 1.39. Among the LLM-fine-tuning baselines, LLaMA3-8B achieves the highest F1 scores. RoBERTa generally exhibits a lower bias score, while BART emerges as a balanced option. Nevertheless, these baselines still lag behind EmoDynamiX by a considerable margin.

\begin{table*}
  \centering
  \scalebox{0.78}{
  \begin{tabular}{lcccccc}
\Xhline{1.25pt}
\multirow{2}{*}{\textbf{Model}} & \multicolumn{3}{c}{ESConv} & \multicolumn{3}{c}{AnnoMI} \\
\cline{2-7} & \(\mathcal{M}\)-F1$\uparrow$ & \(\mathcal{W}\)-F1$\uparrow$ & \(\mathcal{B}\)$\downarrow$ & \(\mathcal{M}\)-F1$\uparrow$ & \(\mathcal{W}\)-F1$\uparrow$ & \(\mathcal{B}\)$\downarrow$ \\
\Xhline{1.25pt}
\rowcolor{gray!20}
    \textit{\textbf{EmoDynamiX}} & \textbf{27.70} &  \textbf{32.71} & \textbf{0.45} & \textbf{27.92} &  \textbf{35.33} & \textbf{0.50}\\
    w/o Graph Learning & 25.72\color{red}$_{\downarrow 1.98}$ & 29.31\color{red}$_{\downarrow 3.40}$ & 0.78\color{red}$_{\uparrow 0.33}$ & 26.95\color{red}$_{\downarrow 0.97}$ & 29.46\color{red}$_{\downarrow 5.87}$ & 0.73\color{red}$_{\uparrow 0.23}$ \\
    w/o Mixed Emotion & 25.90\color{red}$_{\downarrow 1.80}$ & 29.45\color{red}$_{\downarrow 3.26}$ & 0.66\color{red}$_{\uparrow 0.21}$ & 24.71\color{red}$_{\downarrow 3.21}$ & 30.25\color{red}$_{\downarrow 5.08}$ & 0.70\color{red}$_{\uparrow 0.20}$ \\
    w/o Discourse Parser & \underline{26.64}\color{red}$_{\downarrow 1.06}$ & \underline{30.12}\color{red}$_{\downarrow 2.59}$ & \underline{0.59}\color{red}$_{\uparrow 0.14}$ & \underline{27.04}\color{red}$_{\downarrow 0.88}$ & \underline{31.59}\color{red}$_{\downarrow 3.74}$ & \underline{0.60}\color{red}$_{\uparrow 0.10}$ \\
    w/o Dummy Node & 25.46\color{red}$_{\downarrow 2.24}$ & 29.80\color{red}$_{\downarrow 2.91}$ & 0.73\color{red}$_{\uparrow 0.28}$& 24.73\color{red}$_{\downarrow 3.19}$ & 29.00\color{red}$_{\downarrow 6.33}$& 0.72\color{red}$_{\uparrow 0.22}$ \\
\Xhline{1.25pt}
  \end{tabular}}
  \caption{\label{ablation}
    Evaluation results of ablation study.
  }
\end{table*}

\begin{figure*}[t]
\centering
  \scalebox{0.98}{ % Adjust the scale factor as needed
    \includegraphics[width=\linewidth]{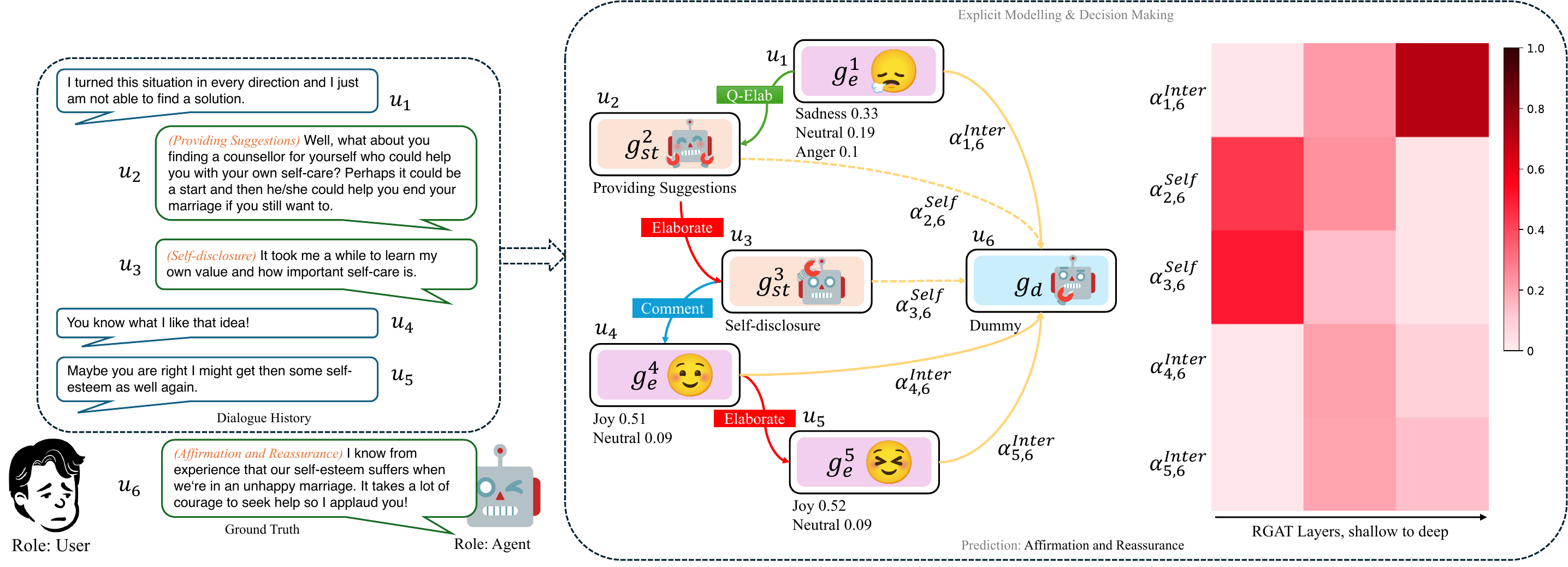}
  }
  \caption{Case study: Dialogue history and ground truth (left); visualization of the heterogeneous graph structure (middle); attention weights of the dummy node edges (right).}
  \label{fig:case}
\end{figure*}

\subsection{Ablation Study}

We conducted ablation studies (Table~\ref{ablation}) and investigated whether the following modules improves the results of next strategy prediction:

%We conducted ablation studies (Table~\ref{ablation}) around following research questions:
\noindent\textbf{Modelling user emotions and agent strategies in dialogue context.} We compared two simplified versions: (1) flattened dialogue history only (\textit{RoBERTa} in Table~\ref{result-main}); (2) flattened context with emotions and strategies inserted as tags (\textit{w/o Graph}). Although \textit{w/o Graph} outperforms \textit{RoBERTa}, there remains a significant gap compared to EmoDynamiX. This indicates that while incorporating emotions and strategies is beneficial, the effectiveness is still limited without our graph-learning module.

%\noindent\textbf{Is Emotion \& Strategy beneficial for strategy prediction?} We compared two simplified versions: (1) flattened dialogue history only (\textit{RoBERTa} in Table~\ref{result-main}); (2) flattened context with emotions and strategies inserted as tags (\textit{w/o Graph}). Although \textit{w/o Graph} outperforms \textit{RoBERTa}, there remains a significant gap compared to EmoDynamiX. This indicates that while incorporating emotions and strategies is beneficial, the effectiveness is still limited without our graph-learning module.
\noindent\textbf{Modelling mixed emotions.} We modelled user emotional states with one-hot vectors instead (\textit{w/o Mixed Emotion}). The resulting decreases in all metrics highlight the importance of leveraging emotion distributions, not just labels, for capturing fine-grained user emotion states.
%\noindent\textbf{Does Mixed-emotion bring better results?} We modelled user emotional states with one-hot vectors instead (\textit{w/o Mixed Emotion}). The resulting decreases in all metrics highlight the importance of leveraging emotion distributions, not just labels, for capturing fine-grained user emotion states.

\noindent\textbf{Modelling discourse structure.} We connected the nodes in simple sequential order instead of discourse structure (\textit{w/o Discourse Parser}). Although this led to drops in all metrics, the declines were not substantial. We hypothesize that the domain gap between the STAC dataset and the ESC datasets may limit the discourse parsing module's effectiveness.

%\noindent\textbf{Do we need discourse structure?} We connected the nodes in simple sequential order instead of discourse structure (\textit{w/o Discourse Parser}). Although this led to drops in all metrics, the declines were not substantial. We hypothesize that the domain gap between the STAC dataset and the ESC datasets may limit the discourse parsing module's effectiveness.

\noindent\textbf{Use of dummy nodes for information aggregation} We replaced dummy nodes with traditional mean-max pooling  \citep{joshi2021dialograph} (\textit{w/o Dummy Node}). We observed performance decreases, with a more significant decline on AnnoMI, indicating that our dummy node design is particularly beneficial in low-resource settings.
%\noindent\textbf{Are dummy nodes better for information aggregation?} We replaced dummy nodes with traditional mean-max pooling  \citep{joshi2021dialograph} (\textit{w/o Dummy Node}). We observed performance decreases, with a more significant decline on AnnoMI, indicating that our dummy node design is particularly beneficial in low-resource settings.

\section{In-depth Analysis of EmoDynamiX}
\label{sec:case_study}

We illustrated a case study using a snippet from ESConv, as shown in Figure~\ref{fig:case}. The case involves the agent deciding which strategy to apply after the user's emotional state has transitioned positively from \textit{Frustration} (as \textit{Frustration} is not a category in DailyDialog, our mixed-emotion module models it as moderated sadness with a little anger) to \textit{Joy}. The ground truth here is \textit{Affirmation and Reassurance}, which acknowledges the user's positive transition and encourages consolidation of positive mood. By looking at the attention weights of dummy node edges, we can observe the contribution of each node in decision-making. We notice that, as the RGAT layer deepens, the dummy node shifts its attention from previously applied strategies to the user's emotional transition, with higher weights applied to edges connecting with the emotion state nodes (especially \textit{Frustration}). In short, our graph-learning module effectively captures clues from emotion/strategy dynamics.

\begin{figure}[t]
\centering
\scalebox{0.98}{
\includegraphics[width=\columnwidth]{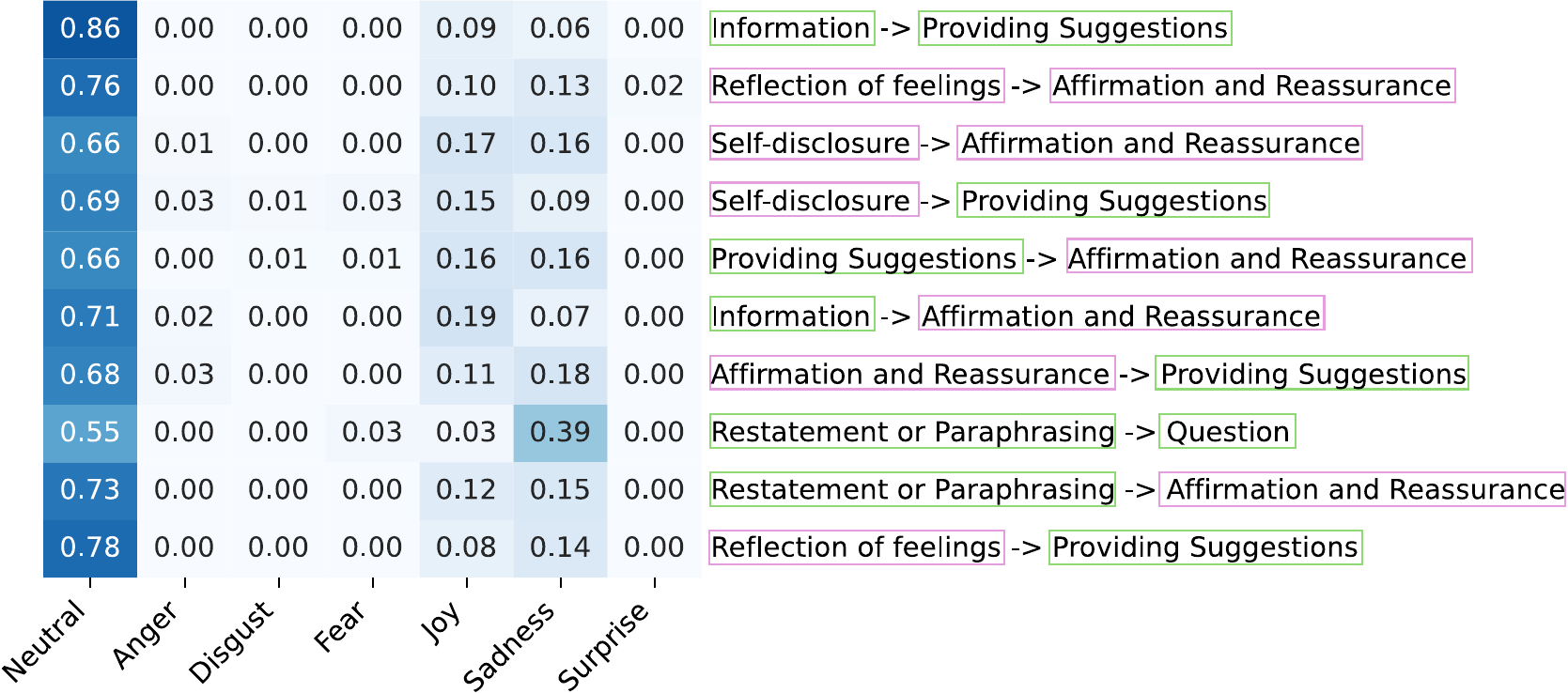}
}
\caption{Analysis on the correlation between the top-10 disagreement patterns (Ground Truth -> Prediction) and their most influential emotion categories.}
\label{fig:vis_error_analysis}
\end{figure}

We further study the disagreements between the predictions and human strategies using the confusion matrix (Appendix~\ref{sec:cm_esconv}). Determining the appropriate timing for \textcolor{violet}{emotion-related} dialogue strategies versus \textcolor{teal}{task-oriented} ones is particularly challenging. The model frequently predicts \textit{Providing Suggestions} whereas the human strategies are among the three emotion-related ones: \textit{Reflection of feelings}, \textit{Self-disclosure}, and \textit{Affirmation and Reassurance}. Notably, over 60\% of emotion-related strategies were categorized as task-oriented ones, highlighting the difficulty in making choices between these two strategy categories, as also discussed by \citet{galland2022adapting}. We further analyze the disagreement patterns between predicted and human strategies. By looking at their correlations with the primary emotion categories of emotion nodes with the highest attention weights (Figure~\ref{fig:vis_error_analysis}), we find that "Neutral" contributes a larger proportion to these disagreements in general compared to its overall representation in the ERC module’s output distribution (59.22\%, as shown in Appendix~\ref{sec:stat_erc_out}). This suggests that strategies are easier to predict when the model can pick up the user emotions expressed in the context.

% Besides, emotions such as "Sadness" could also contribute confusion between content-providing strategies, such as \textit{Restatement or Paraphrasing} -> \textit{Question}.

\section{Conclusions}

In this paper, we propose EmoDynamiX, a socio-emotional dialogue strategy prediction framework that aggregates expert models and uses heterogeneous graphs to model the conversational dynamics of user states and system strategies. Our approach significantly improves all baselines on two public ESC datasets, and takes a step towards transparency by analyzing attention weights in the in-depth study. %We hope that our research can facilitate future work in explicit dialogue decision-making.

\section*{Limitations}

\noindent\textbf{Limitations on Ground Truths} Although the ESC datasets we use have all been evaluated by humans, we cannot fully ensure that no other strategies, aside from the ground truth, could have been effective in the same context. However, we currently lack a protocol for human evaluation at the strategy-planning stage. Besides, human evaluation is more suitable to be performed after the generation stage.

\noindent\textbf{Generalizability to Other Languages}  We evaluated the effectiveness of our proposed architecture using only two English datasets. It remains to be seen whether our approach can generalize to other languages or multi-language settings. It is also worth noting that since EmoDynamiX is based on expert models pre-trained on English datasets to acquire knowledge about discourse structure and emotion recognition, it may inherit cultural biases from these datasets (\citealp{gelfand2011differences}; \citealp{hall1976beyond}), potentially influencing the strategy prediction.

\noindent\textbf{Limitations on Expert Modules}  Since ERC and discourse parsing are not the primary contributions of our research, we did not investigate the impact of using different model architectures or datasets for their training. The training and integration of a cross-domain ERC module and discourse parser could be considered in future studies.

\noindent\textbf{Distance to Practical Application}  Although our method outperformed previous baselines significantly, its performance remains unsatisfactory. This underscores the complexity of the task and indicates that additional work is required to make the socio-emotional strategy predictor a robust component in future ESC agents.

\section*{Ethics Statement}

\textbf{Intent of Technology}  We insist that conversational AI should not be developed to replace humans. Therefore, it is crucial to maintain a clear line between AI and humans \citep{ethique:hal-04702388}. Given that the training data for conversational AI, including the two datasets we selected, are primarily curated by humans, the AI (especially those trained using end-to-end methods) may exhibit human-like behaviors. For instance, in Figure~\ref{fig:task}, the system learns to utilize \textit{Self-disclosure} strategy by expressing feelings of loneliness, which is not consistent with ethical recommendations. We stress that strategies leading to such human-like behaviors should be applied cautiously and potentially restricted in real-world applications to ensure safety. We believe that our approach, which allows us to explicitly set desired behaviors for AI, can provide better control over conversational AI in the future.

\noindent\textbf{Data Privacy}  All experiments were conducted using existing datasets derived from public scientific research. Any personally identifiable and sensitive information, such as user and platform identifiers, has been removed from these datasets. 

\noindent\textbf{Medical Disclaimer}  We do not provide treatment recommendations or diagnostic claims. 

\noindent\textbf{Transparency}  We detail the statistics of the datasets and the hyper-parameter settings of our method. Our analysis aligns with the experimental results. 

\section*{Acknowledgement} 

We thank the anonymous reviewers for their valuable comments. 

This work was partially funded by the ANR-23-CE23-0033-01 SINNet project and benefited from the ANR-23-IACL-0008 project under the France 2030 plan.

% Bibliography entries for the entire Anthology, followed by custom entries
%\bibliography{anthology,custom}
% Custom bibliography entries only
\bibliography{custom}

\clearpage
\appendix

\section{Datasets}
\label{sec:appendix_datasets}

\begin{figure*}[t]
\scalebox{0.94}{
  \includegraphics[width=\linewidth]{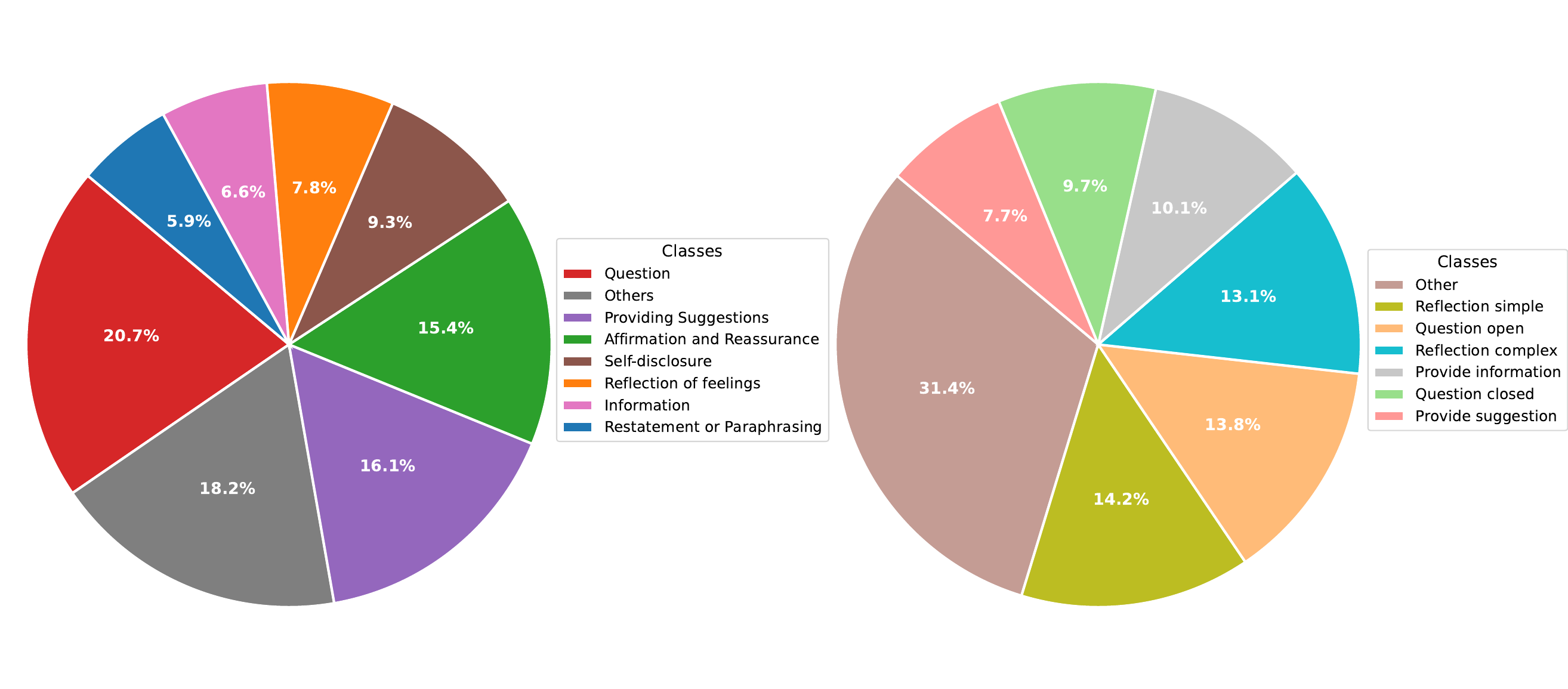}}
  \caption{Strategy distributions of ESConv (left) and AnnoMI (right).}
  \label{fig:dataset_distri}
\end{figure*}

\subsection{ESC Datasets}
\label{sec:appendix_datasets_esc}

Following are the two public ESC datasets we use to evaluate our method.

\noindent\textbf{ESConv} \citep{liu-etal-2021-towards} utilizes eight dialogue strategies: \textit{Reflection of Feelings}, \textit{Self-Disclosure}, \textit{Question}, \textit{Affirmation and Reassurance}, \textit{Providing Suggestions}, \textit{Restatement or Paraphrasing}, \textit{Information}, and \textit{Others}. The distribution of these strategies is depicted in Figure~\ref{fig:dataset_distri}. ESConv collects user feedback scores (ranging from 1 to 5) after every few speaker turns to evaluate the effectiveness of emotional support. Notably, 79.9\% of the scores are above 4 (Good), indicating a high overall quality of emotional support conversations in ESConv, which successfully alleviated users' negative moods. To ensure fair comparisons with previous baselines, we did not perform filtering, though training the strategy predictors on highly rated strategies and using poorly rated ones as negative samples could be beneficial. The top-3 topics include \textit{Ongoing depression}, \textit{Job crisis} and \textit{Break up with parterner}.

\noindent\textbf{AnnoMI} \citep{wu2022anno} categorizes therapist behaviors into 4 high-level types: \textit{Reflection}, \textit{Question}, \textit{Input}, and \textit{Other}. These high-level behaviors are further broken down into 9 fine-grained strategies: \textit{Simple Reflection}, \textit{Complex Reflection}, \textit{Open Question}, \textit{Closed Question}, \textit{Information}, \textit{Advice}, \textit{Giving Options}, \textit{Negotiation/Goal-setting}, and \textit{Other}. Since AnnoMI is a small dataset and has very unbalanced strategy distribution, we merged \textit{Advice}, \textit{Giving Options}, and \textit{Negotiation/Goal-setting} into a single strategy: \textit{Provide Suggestion}, which is aligned with ESConv. The distribution of these strategies is illustrated in Figure~\ref{fig:dataset_distri}. AnnoMI comprises 110 (82.7\%) high-quality dialogues and 23 (17.3\%) low-quality dialogues. To ensure our strategy predictor learns strategies that positively impact users, we excluded all low-quality conversations, retaining only the high-quality ones. The top-3 topics in AnnoMI are \textit{Reducing alcohol consumption}, \textit{Smoking cessation} and \textit{Weight loss}.

\subsection{Datasets for Pre-training Expert Models}
\label{sec:appendix_datasets_expert}

\noindent\textbf{STAC} \citep{asher-etal-2016-discourse} is used for pre-training our discourse parser. It is a multi-party dialogue corpus collected from an online game. It contains 1,081 dialogues, with an average of 8.5 speaker turns per dialogue. STAC includes 16 discourse dependency categories: \textit{Comment, Clarification Question, Elaboration, Acknowledgment, Continuation, Explanation, Conditional, Question-Answer Pair, Alternation, Question-Elaboration, Result, Background, Narration, Correction, Parallel,} and \textit{Contrast}.

\noindent\textbf{DailyDialog} \citep{li-etal-2017-dailydialog} is used to pre-train our emotion recognition module. It is an ERC dataset collected from an English learning website. Its topics are closer to everyday issues and thus better suited for ESC compared to other popular counterparts collected from TV shows (\citealp{poria-etal-2019-meld}; \citealp{zahiri2018emotion}) or actor performances \citep{busso2008iemocap}. DailyDialog includes 13,118 multi-turn dialogues, with an average of 7.9 speaker turns per dialogue. The emotion labels in this dataset encompass Ekman’s six basic emotions (\textit{Anger, Disgust, Fear, Joy, Sadness, Surprise}) and a \textit{Neutral} class.

\section{Definitions of ESC Strategies}

\subsection{Strategies in ESConv}
\label{sec:append_def_st_esconv}

\noindent\textbf{Question}: asking for information related to the problem to help the seeker articulate the issues that they face.

\noindent\textbf{Restatement or Paraphrasing}: a simple, more concise rephrasing of the seeker’s statements that could help them see their situation more clearly.

\noindent\textbf{Reflection of Feelings}: describe the help-seeker’s feelings to show the understanding of the situation and empathy.

\noindent\textbf{Self-disclosure}: share similar experiences or emotions that the supporter has also experienced to express your empathy.

\noindent\textbf{Affirmation and Reassurance}: affirm the help-seeker’s ideas, motivations, and strengths to give reassurance and encouragement.

\noindent\textbf{Providing Suggestions}: provide suggestions about how to get over the tough and change the current situation.

\noindent\textbf{Information}: provide useful information to the help-seeker, for example with data, facts, opinions, resources, or by answering questions.

\noindent\textbf{{Others}}: other support strategies that do not fall into the above categories.

\subsection{Strategies in AnnoMI}
\label{sec:append_def_st_annomi}

\noindent\textbf{Question open}: encourage seekers to elaborate on their thoughts, feelings, and experiences, fostering self-exploration and insight. These questions cannot be answered with a simple yes or no and help build rapport and understanding.

\noindent\textbf{Question closed}: gather specific information, confirm details, or clarify points with concise responses. They are less exploratory but essential for obtaining precise information and ensuring clarity in the conversation.

\noindent\textbf{Reflection simple}: use statements that convey understanding or facilitate seeker-supporter exchanges. Simple reflection conveys understanding of what the seeker has said and adds little extra meaning.

\noindent\textbf{Reflection complex}: use reflective statements that show a deeper understanding of the perspective of the seeker and add substantial meaning or emphasis to what the seeker has said.

\noindent\textbf{Provide suggestion}: provide suggestions (Advice, Options, Goal-Setting) about how to change, but be careful to not overstep and tell them what to do.

\noindent\textbf{Provide information}: provide useful information to the help-seeker, for example with data, facts, opinions, resources, or by answering questions.

\noindent\textbf{{Other}}: exchange pleasantries and use other support strategies that do not fall into the above categories.

\begin{figure}[htbp]
\begin{tcolorbox}[colback=gray!10, colframe=black, title=Prompting Template]
\small
    \textbf{\textit{Task Description}}\\
You are an intelligent emotional support assistant dedicated to helping people cope with stress and depression. To effectively comfort your users, you must select the appropriate dialogue strategy based on the context of the conversation and the user's emotional state. Choose from the following \(x\) types of strategies:\\
\#\# \textit{Strategy Descriptions} \#\#\\
...\\
\textbf{Example 1} \\
\# Dialogue context \# \\
... \\
\# Output \# \\
... \\
\textbf{Example 2} \\
... \\
\textbf{\textit{Dialogue Context}}\\
supporter: (\textit{strategy}) ...\\
seeker: (\textit{emotion: optional})...
\tcblower
\small
\textbf{\textit{Task}}\\
Now select the appropriate dialogue strategy for the next utterance according to the task description and dialogue context above (one answer only, no description), output should be in this format: (\textit{strategy}).
\end{tcolorbox}
\caption{Prompting template for LLMs. Examples are optional.}
  \label{fig:prompt_template}
\end{figure}

\section{Baselines}
\label{sec:baselines}

\textbf{ChatGPT} \citep{chatgpt} is an advanced language model with 175 billion parameters developed by OpenAI. Using reinforcement learning from human feedback (RLHF), it generates human-like text and excels in natural language processing tasks such as conversation and content creation. The prompting is constructed with the template in Figure~\ref{fig:prompt_template} and corresponding strategy definitions in Appendix~\ref{sec:append_def_st_esconv} or Appendix~\ref{sec:append_def_st_annomi}. To facilitate few-shot learning, we built a case library using the training data and extracted two examples from this library for each inference on the test data. We employed example extraction based on similarity scores computed from sentence-BERT \citep{reimers-gurevych-2019-sentence} embeddings.

\noindent\textbf{LLaMA3 (70B \& 8B)} \citep{llama3} is a series of instruction-tuned language models developed by Meta, featuring parameter sizes ranging from 8 to 70 billion. These models are specifically optimized for dialogue use cases and demonstrate superior performance compared to many existing open-source chat models on standard industry benchmarks. For the 70B variant, the prompting templates and few-shot methodologies align with those previously described. In contrast, the 8B variant uses flattened dialogue context with speaker tags as input (same for RoBERTa and BART) and employs LoRA \citep{hu2021lora} for parameter-efficient fine-tuning.

\noindent\textbf{RoBERTa} \citep{liu2019roberta} is a transformer-based language model that enhances BERT by using more training data, larger batch sizes, and dynamic masking, while removing the Next Sentence Prediction objective. Its robust training approach makes it more effective than the original BERT model across multiple benchmarks.

\noindent\textbf{BART} \citep{lewis-etal-2020-bart} is a sequence-to-sequence model that combines a bidirectional encoder and an autoregressive decoder, effectively blending BERT and GPT architectures. It's trained to reconstruct original text from corrupted input, making it highly versatile for tasks like text generation, summarization, and translation.

\noindent\textbf{MISC} \citep{tu-etal-2022-misc} is based on BlenderBot and integrates commonsense knowledge from COMET with a mixed strategy mechanism to simultaneously predict support strategies and generate responses.

\noindent\textbf{MultiESC} \citep{cheng-etal-2022-improving} is a specialized ESC framework based on BART. It features a look-ahead strategy planning mechanism inspired by A* search algorithm to maximize the expected user feedback.

\noindent\textbf{KEMI} \citep{deng-etal-2023-knowledge} is based on BlenderBot and integrates domain-specific case knowledge from HEAL with graph querying. The queries are constructed with commonsense knowledge extracted from COMET. KEMI also simultaneously predict support strategies and generate responses.

\noindent\textbf{TransESC} \citep{zhao-etal-2023-transesc} is a specialized ESC framework built upon BlenderBot. It models dialogue state transitions using a graph-based approach and integrates emotion recognition as an additional training objective, utilizing ground-truth emotion labels predicted by an off-the-shelf ERC model. TransESC's modeling of the user's emotional state also leverages commonsense knowledge from COMET.

\section{Implementation Details}

\subsection{Implementation of the Preference Bias Score}
\label{sec:preference_bias}

\textbf{Preference} $p_i$ indicates the degree to which the model favors strategy $i$ over others. It is calculated iteratively using the confusion matrix according to the following formula:

\matt{\begin{equation}
  p_i' = \frac{\sum_{j}(w_{ij}p_j)/(p_i+p_j)}{\sum_{j}w_{ji}/(p_i+p_j)}
\end{equation}}

Here, $p_i'$ denotes the updated preference for strategy $i$ in the next iteration, and $w_{ij}$ represents the frequency with which the model predicts strategy $i$ when the actual ground truth is strategy $j$. Initially, all preferences $p_i$ are set to 1. In our implementation, we perform 20 iterations of this process.

\textbf{Preference Bias} is the standard deviation of $p$:

\matt{\begin{equation}
  \mathcal{B} = \sqrt{\frac{\sum_{i=1}^N(p_i-\overline{p})^2}{N}}
\end{equation}}

\subsection{Implementation Details for Submodules}
\label{sec:append_impl_sub_modules}

\textbf{Discourse Parser}  We followed exactly the same train/dev/test split and training hyperparameters in the original paper \citep{chi-rudnicky-2022-structured}. The initial learning rate is set to 2e-5 with a linear decay to 0 for 4 epochs. The batch size is 4. The first 10\% of training steps is the warmup stage. We tested the discourse parser on the test set and got a 59.0 F1 on link and relation predictions.

\noindent\textbf{ERC Module}  For the pre-training of the ERC module, we split the DailyDialog dataset as provided in the original repository\footnote{http://yanran.li/dailydialog}. The learning rate was set to 2e-5, with 500 warm-up steps and a weight decay of 1e-3. The model was trained for 12,000 steps, and the best model, determined based on performance on the validation set, was used for inference on the test split. The results on the test set were as follows: an accuracy score of 82.26, a macro F1 score of 53.0, and a weighted F1 score of 83.54.

\subsection{Hyperparameter Settings}
\label{sec:hyps}

For training EmoDynamiX, we configured the batch size to 16 and set the learning rate to 4e-6, with 500 warm-up steps and a weight decay of 1e-3. Additional hyperparameters included a dimensionality of 512 for the heterogeneous graph embeddings, an initial temperature parameter \(\tau\) of 0.5 for the mixed user emotion state module, and 3 layers for relational graph attention. EmoDynamiX was trained for 3000 steps on ESConv and 1200 steps on AnnoMI, separately.

For the pre-training of the ERC module, we set the learning rate to 2e-5, with 500 warm-up steps and a weight decay of 1e-3, training the model for 12,000 steps. For the pre-training of the discourse parser, we adhered to the hyperparameter settings detailed in \citet{chi-rudnicky-2022-structured}. All training procedures were conducted on a single Nvidia GeForce RTX 4090 GPU.

\section{Supplementary Materials for Analysis}

\subsection{Impact of the Initialization of \texorpdfstring{$\tau$}{tau}}
\label{sec:appendix_temperature}

\begin{figure}[t]
  \includegraphics[width=\columnwidth]{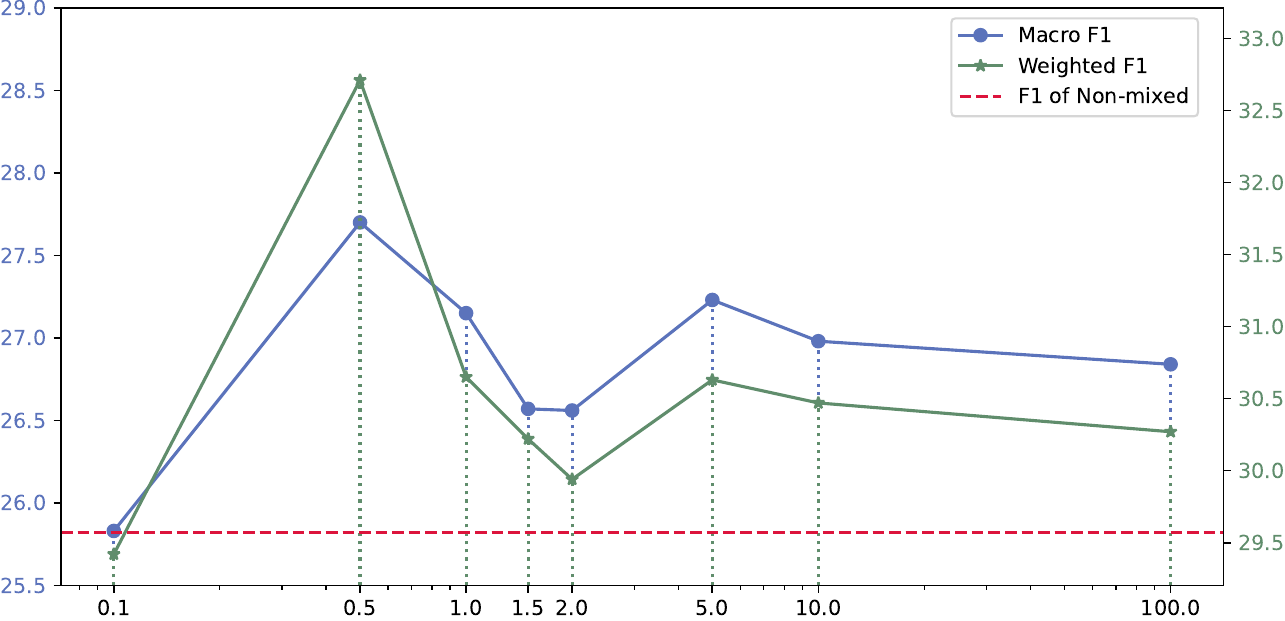}
  \caption{Analysis on the initialized value of $\tau$.}
  \label{fig:t}
\end{figure}

The temperature parameter \(\tau\) adjusts the shape of the probability distributions used in our mixed-emotion module. To understand the impact of the initial value of \(\tau\), we performed extensive experiments, and the results are displayed in Figure~\ref{fig:t}. We observed that while both "sharpening" (lower \(\tau\)) and "softening" (higher \(\tau\)) the distribution can positively impact the overall model performance, outperforming the original distribution (\(\tau=1\)), "sharpening" the distribution makes the best results. This differs slightly from our expectations, as soft probabilities are typically more advantageous in learning paradigms like knowledge distillation \citep{hinton-knowledge-distillation}, which uses distributional knowledge to transfer learning from teacher to student models. We speculate that this outcome might be influenced by the data distribution in the DailyDialog dataset. Since "Neutral" comprises 83\% of the labels, it is usually the largest or second largest category in label distributions. If the label is not "Neutral," "Neutral" plays a significant role, modulating the level of the primary emotion label, such as joy or anger. Conversely, when the label is "Neutral," the second largest category provides additional information, like "Neutral" with a hint of anger or sadness. Our results indicate that emphasizing the primary emotion category while retaining the contribution of the second-highest "moderator" category is beneficial to the overall predictive performance.

We also explored more extreme settings. When \(\tau\) approaches 0, the distribution resembles a one-hot vector, leading to performance similar to the model without the mixed-emotion module (indicated by the red line in the figure). Conversely, initializing \(\tau\) too high (100 or more) over-softens the distributions, allowing minor classes to introduce noise, which results in a decline in performance compared to lower \(\tau\) values.

\subsection{Output Statistics of the ERC Module}
\label{sec:stat_erc_out}

Table~\ref{tab:erc_stat_esconv} presents the output statistics of our pre-trained ERC module on the ESConv test set, alongside a comparison with the original label distribution of DailyDialog. Notably, the dialogue scenes in ESConv exhibit a higher emotional intensity compared to those in DailyDialog.

\begin{table}[h]
    \centering
    \begin{tabular}{rcc}
\Xhline{1.25pt}
         & ESConv & DailyDialog \\
\Xhline{1.25pt}
    Anger & \textbf{1.83} & 0.99 \\
    Disgust & \textbf{0.70} & 0.34 \\
    Fear & \textbf{0.61} & 0.17 \\
    Joy & \textbf{20.17} & 12.51 \\
    Sadness & \textbf{17.17} & 1.12 \\
    Surprise & 0.31 & \textbf{1.77} \\
    Neutral & 59.22 & \textbf{83.10} \\
\Xhline{1.25pt}
    \end{tabular}
    \caption{Comparison between the output label distribution of our ERC module on ESConv and the original label distribution of DailyDialog.}
    \label{tab:erc_stat_esconv}
\end{table}

\subsection{Confusion Matrix of EmoDynamiX on ESConv}
\label{sec:cm_esconv}

\begin{figure*}[t]
  \includegraphics[width=\linewidth]{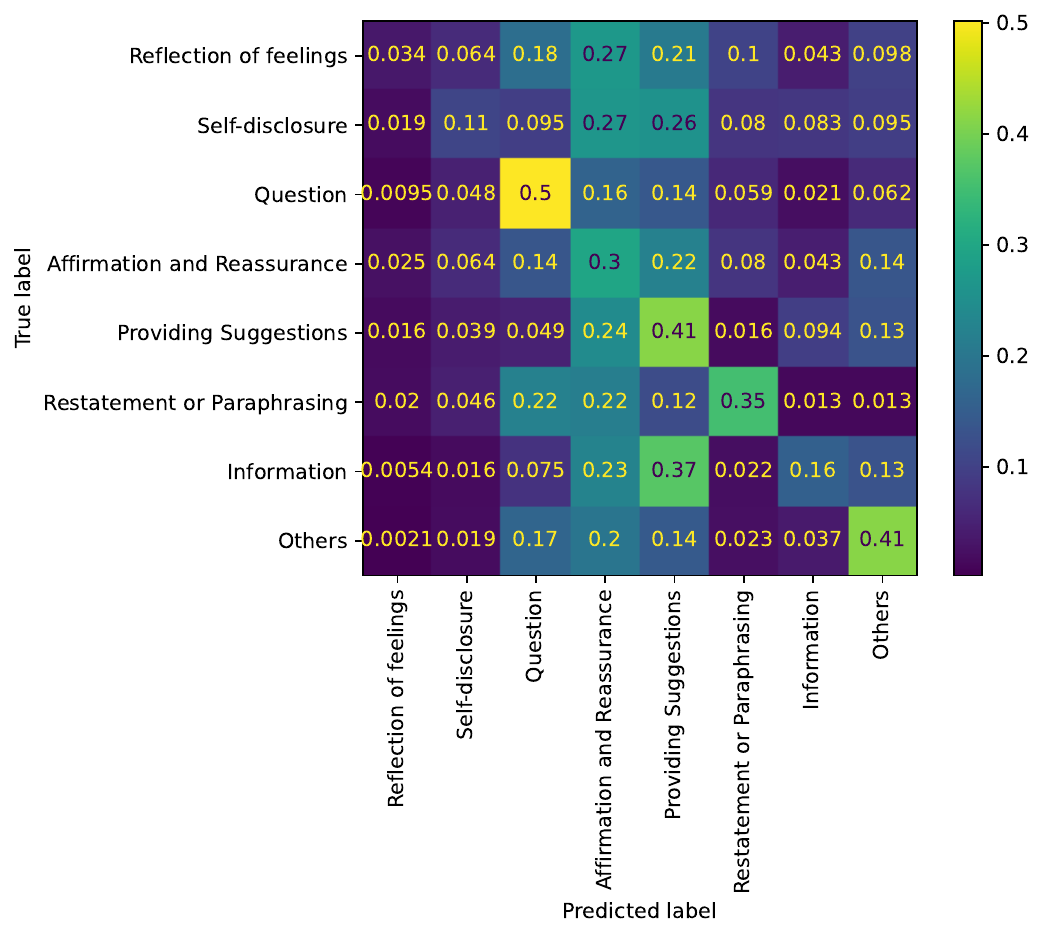}
  \caption{Normalized confusion matrix of EmoDynamiX on the test set of ESConv.}
  \label{fig:cm_esconv}
\end{figure*}

See Figure~\ref{fig:cm_esconv}.

\end{document}